\documentclass[preprint,12pt,3p]{elsarticle}
\usepackage{geometry}
\usepackage{placeins}
\usepackage{xspace}
\usepackage{caption}
\usepackage{subcaption}
\usepackage{graphicx}
\usepackage{url}
\usepackage[hidelinks]{hyperref}
\usepackage{booktabs}
\usepackage{threeparttable}
\usepackage{multirow}
\usepackage{float}
\usepackage{amsmath}

\usepackage{cleveref}
\usepackage{longtable}
\usepackage{natbib}

\usepackage{graphicx}
\usepackage{amssymb}

\usepackage{lineno}
\usepackage[dvipsnames]{xcolor}
\usepackage{ulem}

\definecolor{nyupurple}{RGB}{150, 35, 140}

\journal{arXiv}

\begin{document}

\begin{frontmatter}


\title{Situational-Aware Multi-Graph Convolutional Recurrent Network (SA-MGCRN) for Travel Demand Forecasting During Wildfires}



\author[label1]{Xiaojian Zhang}
\address[label1]{Department of Civil and Coastal Engineering, University of Florida}
\ead{xiaojianzhang@ufl.edu}

\author[label1]{Xilei Zhao}
\ead{xilei.zhao@essie.ufl.edu}

\author[label1]{Yiming Xu}
\ead{yiming.xu@ufl.edu}

\author[label2]{Ruggiero Lovreglio\corref{cor1}}
\address[label2]{School of Built Environment, Massey University}
\ead{r.lovreglio@massey.ac.nz}

\author[label3]{Daniel Nilsson}
\address[label3]{Department of Civil and Natural Resources Engineering, University of Canterbury}
\ead{daniel.nilsson@canterbury.ac.nz}

\cortext[cor1]{Corresponding author. Postal address: School of Built Environment, Massey University, East Precinct Albany Expressway, SH17, Albany, Auckland 0632, New Zealand.}

\begin{abstract}
Real-time forecasting of travel demand during wildfire evacuations is crucial for emergency managers and transportation planners to make timely and better-informed decisions. However, few studies focus on accurate travel demand forecasting in large-scale emergency evacuations. Therefore, this study develops and tests a new methodological framework for modeling trip generation in wildfire evacuations by using (a) large-scale GPS data generated by mobile devices and (b) state-of-the-art AI technologies. The proposed methodology aims at forecasting evacuation trips and other types of trips. Based on the travel demand inferred from the GPS data, we develop a new deep learning model, i.e., Situational-Aware Multi-Graph Convolutional Recurrent Network (SA-MGCRN), along with a model updating scheme to achieve real-time forecasting of travel demand during wildfire evacuations. The proposed methodological framework is tested in this study for a real-world case study: the 2019 Kincade Fire in Sonoma County, CA. The results show that SA-MGCRN significantly outperforms all the selected state-of-the-art benchmarks in terms of prediction performance. Our finding suggests that the most important model components of SA-MGCRN are evacuation order/warning information, proximity to fire, and population change, which are consistent with behavioral theories and empirical findings.

\end{abstract}


\end{frontmatter}



\section{Introduction}
\label{S:1}

Wildfires are posing a growing threat to communities across the U.S. and worldwide. Studies have found that the consequences of wildfires have been aggravated in recent years, mainly due to accelerating climate change risks and rapid expansion of the wildland-urban interface (WUI) \citep{kuligowski2020modelling,zhao2022estimating,radeloff2018rapid}. For example, in recent years, Sonoma County in California experienced multiple large wildfires (e.g., the 2017 Tubbs Fire, the 2019 Kincade Fire, and the 2020 Glass Fire), which forced thousands of people to evacuate nearly every year. Wildfires are also becoming a threat in winter season as demonstrated by the 2021 Marshall Fire. In terms of structures lost, the Marshall Fire (December 30, 2021 – January 1, 2022) is the most destructive fire in Colorado history \citep{COSun}. To reduce the life safety risks of wildfires and enhance the resilience of WUI communities, it is crucial to improve real-time decision support for emergency managers by accurately forecasting the \textit{travel demand} (including evacuation trips and other types of trips) during wildfire evacuations.

Significant research gaps exist in terms of travel demand forecasting in large-scale emergency evacuations. To date, evacuation demand (i.e., time-dependent evacuation trip generation) is generally estimated by using survey data and logistic regression \citep{murray2013evacuation}, both of which have some limitations. Specifically, despite generating valuable insights on household evacuation behavior, survey data have relatively small sample sizes (e.g., only hundreds of data points) and provide a low-resolution timeline (e.g. 2–6 hour resolution) of household decisions during the evacuation \citep{zhao2022estimating,lovreglio2020calibrating,fu2007modeling}. Based on the survey data, logistic regression is commonly used to model the binary evacuation decision (to evacuate or not) of households \citep{murray2013evacuation,kuligowski2020modelling}, but logistic regression cannot capture the potential nonlinearities and interactions between the evacuation decision and independent variables and thus often lead to inaccurate predictions \citep{zhao2020modelling}. In addition to trips for evacuation purposes, other types of trips, such as background trips and intermediate trips, play essential roles in measuring transportation network performance and estimating travel time \citep{murray2013evacuation,zeigler1981evacuation}. However, nearly no prior work has estimated the total amount of trips generated during evacuations \citep{mcghee2006operational,murray2013evacuation}. 

To tackle these research gaps, we aim to develop a new trip generation model including evacuation trips and other types of trips for a wildfire evacuation by using GPS data and AI technologies. GPS data typically contain millions of location signals from mobile devices such as smartphones and smartwatches. As such, GPS data provides highly granular spatiotemporal trajectories of people’s movements during an evacuation \citep{zhao2022estimating}. GPS data has shown great potential for estimating and understanding evacuation behavior for different types of disasters \citep{horanont2013large,yabe2020effects,zhao2022estimating}. With the GPS data, hourly trip generation at the zonal level can be inferred, which can then be used to model travel demand. 

With the inferred trips, we develop a new deep learning model, a \textbf{S}ituational-\textbf{A}ware \textbf{M}ultiple \textbf{G}raph \textbf{C}onvolutional \textbf{R}ecurrent \textbf{N}etwork (\textbf{SA-MGCRN}), along with a model updating scheme to achieve real-time forecasting of travel demand during wildfire evacuations. The proposed model uses a graph convolutional network (GCN) based on environmental similarity graph, and demographic similarity graph to attach spatial dependency to temporal inputs (e.g., historical travel demand). The output of GCN is subsequently processed with the weather information, evacuation order/warning, fire progression, weekend indicator (i.e., a binary variable indicating if it is Sat./Sun.) and historical travel demand information by the Gated Recurrent Unit (GRU) to capture temporal dependency. Then, a fully-connected layer is used to generate the final prediction for the hourly trip generation at the zonal (i.e., census tract) level. Given the delivery delay issues (which is common among GPS providers) of GPS data, we propose a new model updating scheme to ensure the situational awareness of the AI model. The proposed model and the model updating scheme are empirically evaluated by a real-world case study of Kincade Fire, Sonoma County, CA. The results of this study can be directly used by emergency managers to support real-time decision-making in wildfire evacuations. 

The remaining paper is structured as follows: Section \ref{LR} reviews the related prior work. Section \ref{research design} introduces the methodology that includes three major parts, i.e., trip generation inference, travel demand forecasting, and model updating scheme. Section \ref{data} describes the case study and Section \ref{results} presents the results of the case study. Section \ref{discussion and conclusion} discusses the key findings of the research and concludes the paper.

\section{Literature Review}
\label{LR}

In this section we provide an overview of existing AI applications for evacuation modeling (Section \ref{sub2.1}) and the key actors affecting evacuation wildfire decision-making (Section \ref{sub2.2}).

\subsection{AI applications in evacuation modeling}
\label{sub2.1}
Machine learning techniques provide researchers powerful tools to model human behaviors in both normal conditions \citep{yan2020using, zhao2020prediction, shao2020threshold, cheng2019applying, xu2021identifying} and emergencies such as wildfires and hurricanes \citep{zhao2020modelling, zhao2021using, zhaoamodeling, wang2019extracting, lo2009artificial}. Machine learning method usually generates much more accurate predictions and potentially offers more intuitive interpretations compared to the traditional statistical models, due to their flexible modeling structures that can automatically capture the underlying complex relationships within the data \citep{zhao2020modelling, zhao2020prediction, xu2021identifying}. For example, \citet{zhao2020modelling} applied random forest (a tree-based machine learning model) to model individuals’ pre-evacuation decisions in building fires. The results showed strong nonlinear relationships between key factors (such as the time elapsed after the alarm has started and the decision maker’s personal group size) and building occupants' decisions which were not identified with traditional statistics in previous studies \citep{lovreglio2015model}. More recently, \citet{zhao2021using} has applied random forest to model and explain individual-level evacuation decisions for wildfires and the random forest model showed superior results than the traditional statistical model. 

Recently, deep learning has gained wide popularity in emergency management studies such as traffic speed  and volume forecasting during emergency events, as it can produce highly accurate real-time predictions \citep{rahman2018short, roy2021predicting, huang2021systematic}. For example, \citet{rahman2018short} applied a long short-term memory (LSTM) model to predict traffic speeds in freeways during hurricane Irma's evacuation. \citet{roy2021predicting} used a LSTM model to forecast the traffic volumes of interstate highways during hurricane Matthew and Irma evacuation. \citet{nguyen2019forecasting} developed a deep learning model that consists of two LSTM models to forecasting people’s needs during hurricane Sandy, Harvey, and Irma using social media and weather data. \citet{li2022identifying} adopted a LSTM model and a Bidirectional Encoder Representations from Transformers (BERT) model to identify people's evacuation intent during hurricane Irma using the social media data. However, no work has applied deep learning methods to predict real-time wildfire evacuation demand. Such information may serve as essential inputs for emergency managers to make better-informed decisions during the event.

\subsection{Factors associated with evacuation decision-making}
\label{sub2.2}
The householders’ decision to evacuate or stay during a wildfire has been widely investigated in the literature using different qualitative and quantitative approaches, such as individual interviews, focus group interviews, and questionnaires \citep{kuligowski2021evacuation}. These tools have been used to investigate several wildfire disasters in multiple countries all around the world, revealing factors which can affect the householders’ decision-making. Lists of these factors have been provided in several works. For instance, \citet{folk2019provisional} identified several of these factors and how they affected different components of the decision-making process using the Protective Action Decision Model Framework introduced by \citet{lindell2003communicating}. This list was further expanded by \citet{lovreglio2019modelling, lovreglio2020calibrating, kuligowski2022modeling} to estimate regression models predicting the decision to evacuate as well as householders’ evacuation behavior using data collected after the 2019 Kincade Fire (USA), the 2016 Chimney Tops 2 (USA) and the 2014 bushfire in Parkerville, Stoneville and Mt Helena in the Perth Hills (Australia) and the 2015 bushfire in Adelaide Hills (Australia).

The literature shows that there are multiple internal factors (e.g., demographic, perspectives, memories of previous experiences and attitudes) and external factors (e.g., disaster alert and warning as well as cues from the physical and social environment) affecting protective action decision-making \citep{lovreglio2019modelling}. While focusing on external factors, several researchers have reported that the type and number of warnings and cues affect the householders’ decision to evacuate. For instance, it has been seen that the likelihood of households evacuating increases if they receive mandatory or voluntary evacuation orders \citep{kuligowski2022modeling, lovreglio2019modelling, mclennan2011issues, mozumder2008investigating, strawderman2012reverse, whittaker2016gendered, wong2022understanding}. This trend increases when these orders are given by trusted sources \citep{kuligowski2020modelling}. Furthermore, visible smoke, embers, flames, and information of the fire’s location also increase the number of householders who decide to evacuate \citep{mccaffrey2018should, mclennan2012householder}. However, \citet{lovreglio2020calibrating} did not find significant evidence that the distance of the fire has impact of the decision to evacuate. However, the author claims that this could be linked to the level of uncertainty of their data as they did not have access to the precise home location of the householders as they used their postcode a proxy to infer distance.

When focusing on internal factors, there is an extended list of factors which have been linked to the householders’ decision to evacuate or stay. For instance, several factors (such as gender, age, previous experience, household composition, and having pets and livestock) have proven to affect the decision to evacuate in different ways \citep{alsnih2005understanding, eriksen2010gendered, kuligowski2020modelling, mozumder2008investigating, paveglio2014understanding, whittaker2016gendered, wong2022understanding,strawderman2012reverse}. These variables are easy to be investigated using traditional survey studies and be integrated into evacuation planning models. However, it can be very challenging to collect these individual-level data from a large number of householders in real-time to assist in emergency response.

\section{Approach}
\label{research design}

\begin{figure}[!ht]
  \centering
  \includegraphics[width=0.98\textwidth]{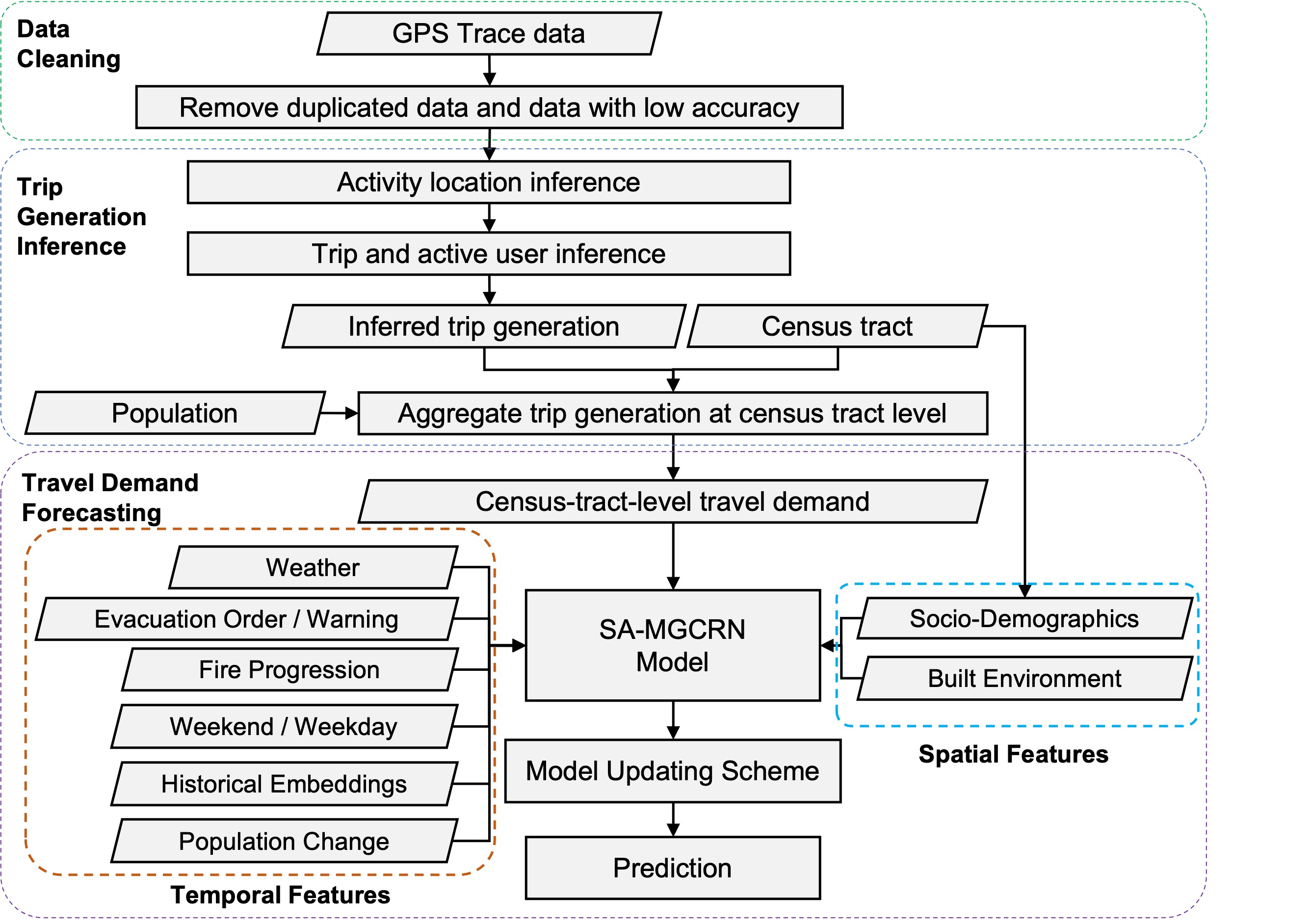}
  \caption{Overall Methodological Framework}\label{fig:overall_frmk}
\end{figure}

This section introduces the process of trip generation inference using GPS data, the travel demand forecasting model based on deep learning, and the model updating scheme. The overall framework of this study is presented in Figure \ref{fig:overall_frmk}. We first clean the GPS trace data by removing duplicated data points and inaccurate data points\footnote{As GPS records usually have spatial measurement errors \citep{zhang2016spatial}, some data providers such as Gravy Analytics would label the accuracy of the latitude and longitude of a GPS record. Modelers can choose an error threshold to filter out highly inaccurate data points. Note that this step can be skipped if the data provider does not provide this data field.}. Then, we implement a trip generation inference algorithm to extract census-tract-level active users and trip generations based on the GPS trace data. After that, we develop a Situational-Aware Multi-Graph Convolutional Recurrent Network (SA-MGCRN) model to forecast the travel demand at the census-tract level. The input of the SA-MGCRN model includes historical census-tract-level travel demand, temporal features (i.e., weather, evacuation order/warning, fire progression, day of the week, historical demand information, and population change), and spatial features (i.e., socio-demographics and built environment). The proposed model will be updated regularly (e.g, daily) to quickly adapt to the fast-changing situations during wildfires. The details of the trip generation inference algorithm, the SA-MGCRN model, and the model updating scheme are explained in the following subsections.

\subsection{Trip generation inference}
\label{trip inference}
\begin{figure}[!ht]
  \centering
  \includegraphics[width=0.99\textwidth]{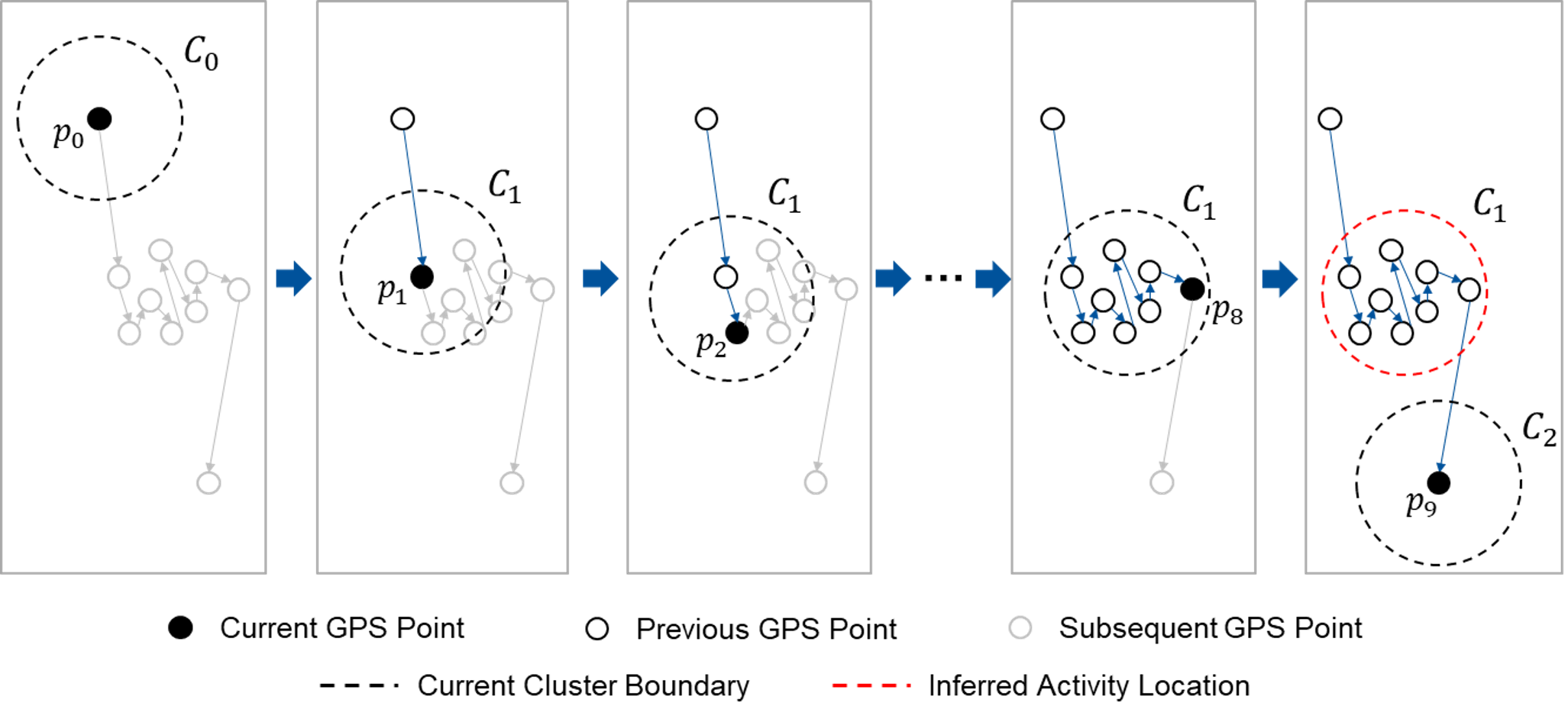}
  \caption{An example to illustrate the incremental clustering method to infer activity locations} \label{fig:clustering}
\end{figure}

In this study, we assume that if a GPS device user stayed at a place for a period of time, the user carried out a certain activity at that place. Based on this assumption, we can extract activity locations by clustering using GPS trace data. This study applied an \textbf{incremental clustering method} \citep{wang2018data, alexander2015origin} to infer activity locations. Incremental clustering is a stream clustering approach that processes the data points in a sequence successively. Given a sample of GPS trace data $P=[p_0,p_1,\dots,p_n]$, the clustering process starts from the first data point $p_0$. We first create a new cluster $C_0$ centering at $p_0$ with radius of $R$. Then we check if the next data point $p_1$ is in cluster $C_0$ by calculating the distance $d_1$ between $p_1$ and center of $C_0$. If $d_1<R$, we aggregate $p_1$ to cluster $C_0$, and update the center of $C_0$. Otherwise, we create another cluster $C_1$ centering at $p_1$ with radius of $R$. We perform the same procedure for the data points in the GPS trace data sample $P$ successively. After that, we check the time duration $t_i$ of each cluster $C_i$ (i.e., time duration that the GPS device user spends in cluster $C_i$). If $t_i$ is no less than a given threshold $T_c$, the cluster $C_i$ is inferred as an activity location. Note that we need to set proper values for cluster radius $R$ and the time threshold $T_c$ to obtain reliable inferred results. In this study, we set the cluster radius $R$ and the time threshold $T_c$ to 500 meters and 5 minutes, respectively \citep{chen2014traces, wang2019extracting}. The activity location inference process is illustrated in Figure~\ref{fig:clustering}. 

We assume that there is a trip between every two consecutive activity locations in a GPS trace trajectory. Based on this assumption and the inferred activity locations, we can further infer trips by linking time-adjacent activity locations, and thus obtain trip origins and destinations. We further aggregate the trip origins to calculate the hourly census-tract-level trip generations for each day. Since each activity corresponds to a unique ID, we also infer the number of active users on a census-tract-level basis for each timestamp. We denote the inferred (from GPS data) trip generations and the number of active users at hour $i$ for day $j$ in census tract $k$ as $M_{ijk}^{GPS}$ and $U_{ijk}^{GPS}$. It is important to note that GPS data is generated by only a small portion of the population and therefore cannot accurately represent overall travel demand at the census-tract level for the whole population. To address this issue, we use the following procedure.

First, we calculate the average number of active users \textit{per day} during all workdays prior to the wildfire for each census tract $k$, denoted by $U_{k}^{GPS}$. This value is used to represent the overall number of active users for each census tract. Mathematically, it can be written as:
\begin{equation}
    U_{k}^{GPS} = \frac{1}{\left| S \right|}\sum_{i}\sum_{j}U_{ijk}^{GPS}, j \in S
\end{equation}
\noindent where  $S$ refers to the set of workdays prior to the wildfire and correspondingly, $\left| S \right|$ refers to the number of workdays (i.e., the size of the set $S$).

Next, we divide the inferred trip generations (i.e., $M_{ijk}^{GPS}$) by the assumed overall number of active users per day (i.e., $U_{k}^{GPS}$) to obtain the inferred trip generation rate per person for each census tract at every timestamp. Specifically,

\begin{equation}
    R_{ijk}^{GPS} = M_{ijk}^{GPS} / U_{k}^{GPS}
\end{equation}

\noindent where the inferred trip generation rate per person at hour $i$ for day $j$ in census tract $k$ is denoted by $R_{ijk}^{GPS}$. Finally, we multiply the trip generation rate per person by the number of population in each census tract (denoted by $P_k$) to estimate the travel demand, or trip generations, of all populations. More formally, trip generations for the whole population at hour $i$ for day $j$ in census tract $k$ is calculated as:

\begin{equation}
    M_{ijk}^{All} = R_{ijk}^{GPS} \times P_k
\end{equation}

By using this procedure, we can obtain a more accurate estimation of travel demand at the census-tract level for the whole population while taking into account the limitations of GPS data.

\subsection{Travel demand forecasting}

Formally, the travel demand forecasting problem in this study can be formulated as follows.\\

\noindent\textbf{\textit{Problem}}: Given a historical travel demand (i.e., evacuation, background and intermediate trips) matrix $\textbf{X}_t$, a temporal feature (such as weather conditions) matrix $\textbf{C}_t$, and a spatial correlation (such as build environment similarity) graph $G$, learn a function $f:\mathbb{R}^{N \times T} \rightarrow \mathbb{R}^{N\times M}$ that maps historical travel demand of all census tracts to the travel demand in next $M$ time intervals:

\vspace{-10pt}
\begin{equation}
\textbf{Y}_t = [X_{t+1},\dots,X_{t+M}] = f(\textbf{X}_t,\textbf{C}_t,G)
\end{equation}
where $N$ is the number of census tracts, and $T$ is time sequence length of input historical travel demand data.

The historical travel demand matrix $\textbf{X}^{N \times T}_t$ represents historical travel demand for all the $N$ census tracts from time $t-T+1$ to time $t$. $\textbf{X}_t = [X_{t-T+1},\dots,X_t]$ where $X_t=[x_t^1,x_t^2,...,x_t^N]$ denote the travel demand of all census tracts at time $t$, and $x_t^i$ denote the travel demand of census tract $i$ at time $t$.

The temporal feature matrix $\textbf{C}^{K \times T}_t$ consists of several temporal features from time $t-T+1$ to time $t$, including weather condition, evacuation order/warning in each census tract, fire progression, weekend indicator and historical travel demand information, where $K$ is the number of temporal features.

The spatial correlation graph $G=(V,E)$ describes the spatial and property correlation between areas. This graph is fused by two graphs to represent environmental and demographic similarities between the areas. In graph $G=(V,E)$, $V=\{v_1,v_2,...,v_N\}$ is a set of nodes (i.e., census tracts). $E$ is a set of edges. Note that the graph $G$ is an undirected graph. If two nodes in $G$ are correlated (e.g., functionally similar), there is an edge between these two nodes. An adjacency matrix $A \in \mathbb{R}^{N \times N}$ can be used to represent the graph $G$. The element $a_{ij} \in A$ is 1 if there is an edge between node $i$ and node $j$; otherwise, 0.

\subsubsection{Overview of model architecture}

\begin{figure}[H]
  \centering
  \includegraphics[width=0.99\textwidth]{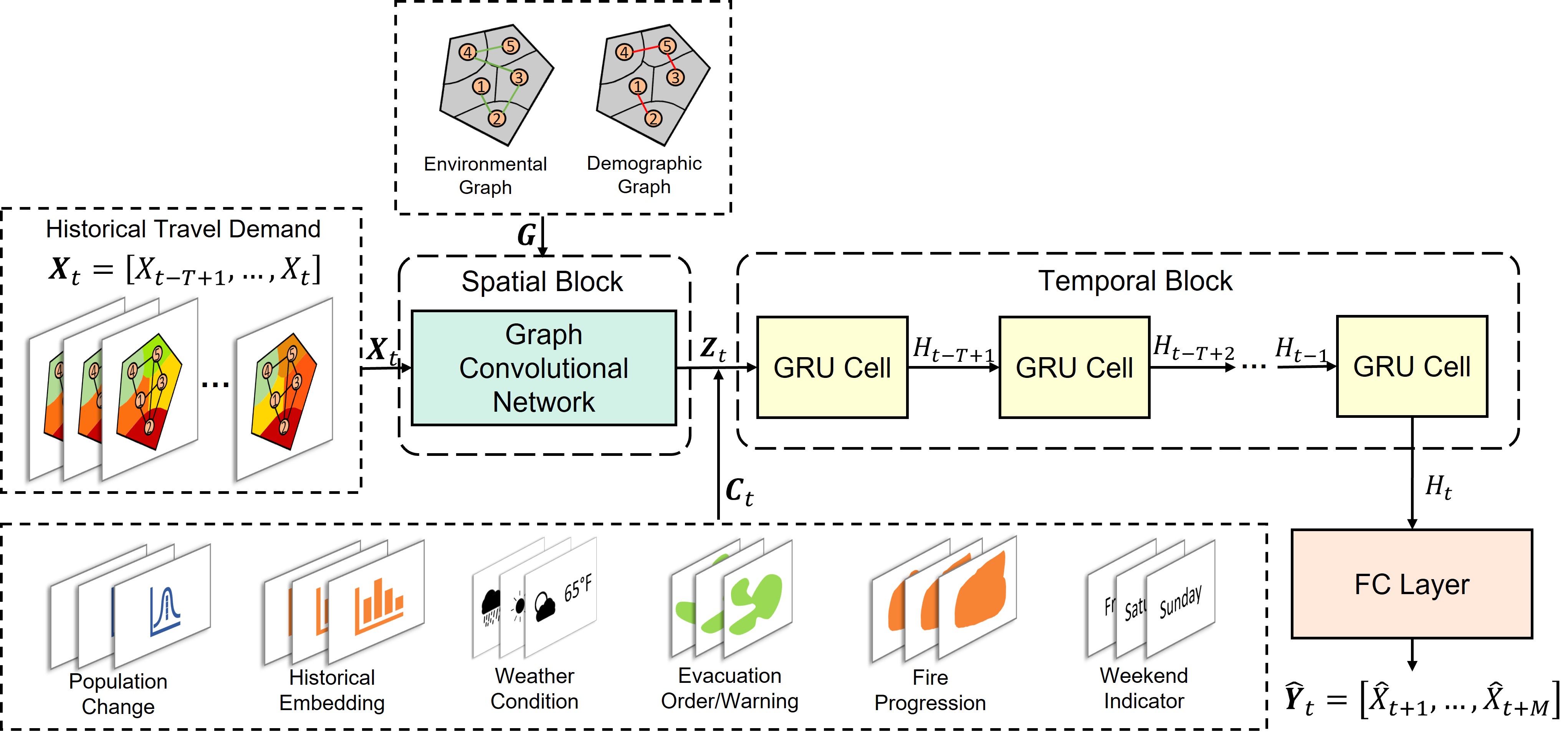}
  \caption{Architecture of the SA-MGCRN model} \label{fig:pred_frmk}
\end{figure}

We propose a \textbf{S}ituational-\textbf{A}ware \textbf{M}ultiple \textbf{G}raph \textbf{C}onvolutional \textbf{R}ecurrent \textbf{N}etwork (\textbf{SA-MGCRN}) model to solve the travel demand forecasting problem. The model architecture is presented in Fig. \ref{fig:pred_frmk}. The proposed SA-MGCRN model is composed of a spatial block, a temporal block, and a fully connected (FC) layer. The spatial block is a graph convolutional network (GCN) based on the spatial correlation graph $G$. The graph $G$ is fused by three graphs including adjacency graph, functional graph, and demographic graph. The GCN takes historical travel demand $\textbf{X}_t$ as input, and generates output $\textbf{Z}_t$ by convolutional computation. The output $\textbf{Z}_t$ is then concatenated with temporal feature matrix $\textbf{C}_t$. After that, the temporal block uses a sequence of gated recurrent unit (GRU) cells to processes $[\textbf{Z}_t,\textbf{C}_t]$ successively. The output of temporal block $H_t$ is subsequently processed by a FC layer to generate the prediction $\widehat{\textbf{Y}}_t = [\widehat{X}_{t+1}, \dots, \widehat{X}_{t+M}]$. The details of the GCN model and the GRU model are introduced in the following subsections.

\subsubsection{Graph Convolutional Network (GCN)}

Travelers in areas with similar properties are more likely to have similar travel behavior \citep{xu2021identifying, zhang2022examining}. Therefore, spatial dependencies are essential to travel behavior prediction \citep{zhao2019t, zhang2022machine, xu2023real}. Graph convolutional network (GCN) is a widely-used model to capture spatial dependencies. GCN performs convolutional operations based on a graph, thus, it can handle graph-structured data \citep{zhao2019t}. Since the census tracts do not have a regular spatial structure but can be represented by a graph, we use GCN to capture spatial dependencies in this study.

Given an adjacency matrix $A$ and the historical travel demand $\textbf{X}_t$, GCN performs convolutional operations using a filter in the Fourier domain. The filter is applied on each node of the graph, thus capturing spatial dependencies between the node and its adjacent nodes. The GCN model is constructed by multiple convolutional layers:

\begin{linenomath}
\begin{equation}
H^{l+1}=\sigma (\tilde{D}^{-\frac{1}{2}} \tilde{A} \tilde{D}^{-\frac{1}{2}} H^l W^l)
\end{equation}
\end{linenomath}\\
\noindent where $H^l$ is the output of layer $l$ and $H^0=\textbf{X}_t$, $\tilde{A}=A+I$ is the adjacency matrix of the graph $G$ with self-connections, $I$ is the identity matrix, $\tilde{D}$ is the diagonal node degree matrix of $\tilde{A}$, and $W^l$ is a layer-specific trainable matrix. $\sigma(\cdot)$ denotes an activation function, such as the $ReLU(\cdot)=max(0,\cdot)$ \citep{nair2010rectified}. A two-layer GCN model \citep{kipf2016semi} is used in this study to capture spatial dependencies. We first calculate $\widehat{A}=\tilde{D}^{-\frac{1}{2}} \tilde{A} \tilde{D}^{-\frac{1}{2}}$ in a pre-processing step. The forward model then takes the form:

\begin{linenomath}
\begin{equation} \label{eq:z}
Z=f(X,A)=softmax(\widehat{A}~ReLU(\widehat{A}XW^0)~W^1)
\end{equation}
\end{linenomath}\\
\noindent where $W^0 \in \mathbb{R}^{C \times H}$ is the input-to-hidden weight matrix, $C$ is the number of input channels (i.e., a $C$-dimensional feature vector for each node), $H$ is the number of hidden units, $W^1 \in \mathbb{R}^{H \times F}$ is the hidden-to-output weight matrix, $F$ is the number of filters, $Z \in \mathbb{R}^{N \times F}$ is the output convolved matrix, and $N$ is the number of nodes. The softmax activation function, defined as $softmax(x_i)=\frac{1}{z}exp(x_i)$ with $z=\sum_i exp(x_i)$, is applied row-wise.

\subsubsection{Gated Recurrent Unit (GRU)}

Gated recurrent unit (GRU) \citep{cho2014properties} is a developed RNN model to process sequence data such as time series and speech signal data. Using the gating mechanism to memorize long-term information, GRU can well deal with the vanishing gradient problem that may happen in traditional RNN models. Compared with other RNN models such as LSTM \citep{hochreiter1997long}, GRU is faster to compute but still offers comparable performance in prediction \citep{chung2014empirical}. Therefore, we use GRU to perform temporal prediction in this study.

\begin{figure}[hbt!]
    \centering
    \includegraphics[width=0.75\textwidth]{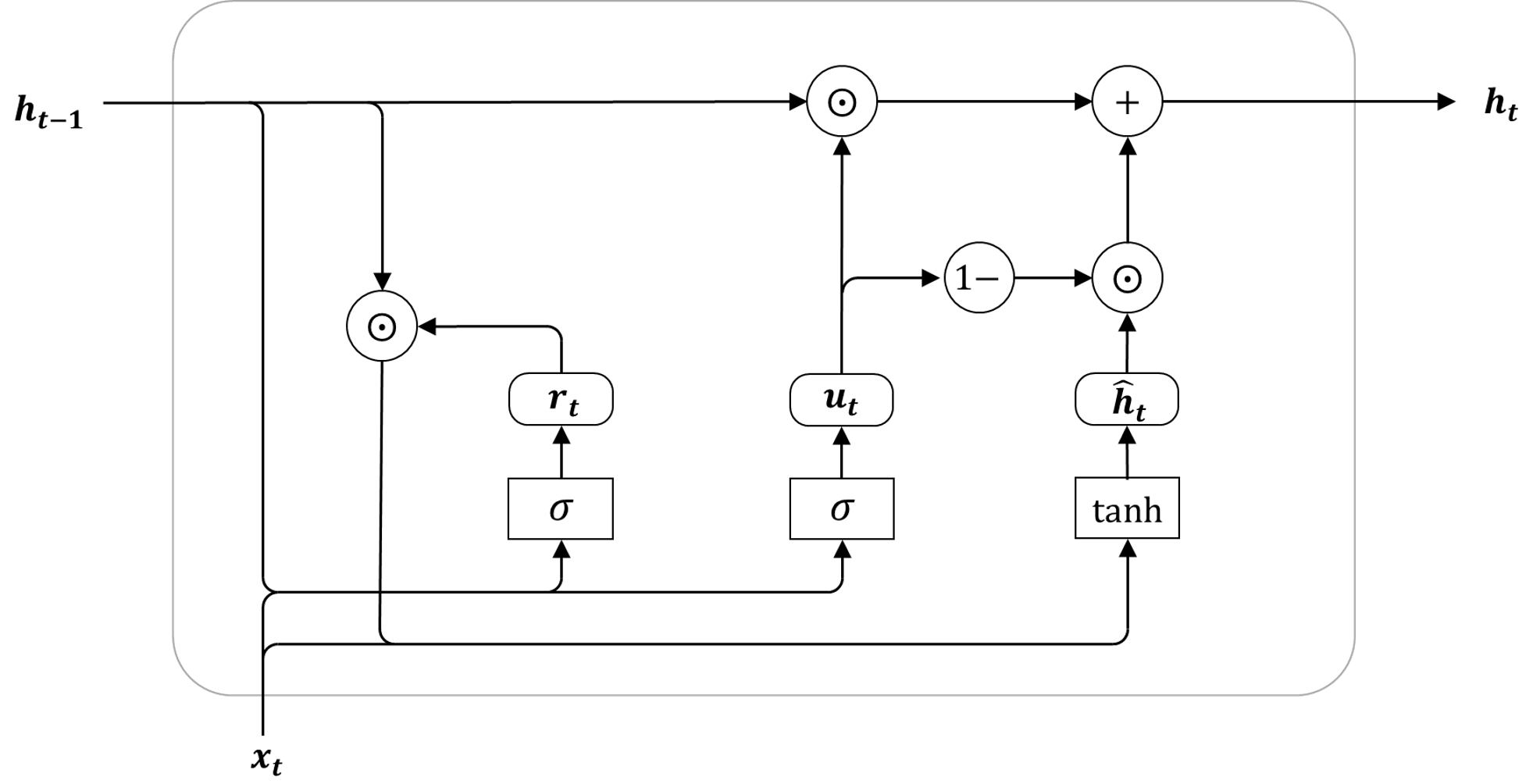}
    \caption{Structure of a GRU cell. $h_{t-1}$ is the previous hidden state. $x_t$ is the current input. $r_t$ is the reset gate and $u_t$ is the update gate. $\hat{h}_t$ is the candidate hidden state and $h_t$ is the current hidden state. $\odot$ is the element-wise product operator. $\sigma$ is the sigmod activation and $tanh$ is the tangent hyperbolic activation. }
    \label{fig:gru}
\end{figure}

GRU uses two gates, reset gate and update gate, to determine what information should be kept and passed to the output. The reset gate controls how much of the previous state information to remember, and the update gate determines how much of the past information needs to be passed to the new state. The structure of a GRU cell is presented in Figure~\ref{fig:gru}. 

The GRU cell takes current input $x_t$ and the previous hidden state $h_{t-1}$ as inputs, and outputs the new hidden state $h_t$. Given $x_t$ and $h_{t-1}$, the reset gates $r_t$ and the update gate $u_t$ are calculated by fully connected layers with the sigmoid \citep{nwankpa2018activation} activation function $\sigma$. Mathematically, for a given time step $t$, the reset gates $r_t$ and the update gate $u_t$ are computed by:

\begin{linenomath}
\begin{equation}
r_t=\sigma(x_tW_{xr}+h_{t-1}W_{hr}+b_r)
\end{equation}
\begin{equation}
u_t=\sigma(x_tW_{xu}+h_{t-1}W_{hu}+b_u)
\end{equation}
\end{linenomath}\\
\noindent where $W_{xr}$, $W_{hr}$, $W_{xu}$, $W_{hu}$ are weight parameters and $b_r$, $b_u$ are biases.

Then we integrate the reset gate $r_t$ with the previous hidden state $h_{t-1}$ and current input $x_t$ to generate the candidate hidden state $\hat{h}_t$ :

\begin{linenomath}
\begin{equation}
\hat{h}_t=tanh(x_tW_{xh}+(r_t \odot h_{t-1})W_{hh}+b_h)
\end{equation}
\end{linenomath}\\
\noindent where $W_{xh}$, $W_{hh}$ are weight parameters, $b_h$ is the bias, and $\odot$ is the element-wise product operator. 

We use the update gate $u_t$, the candidate hidden state $\hat{h}_t$, and the previous hidden state $h_{t-1}$ to compute the current hidden state $h_t$. The hidden state $h_t$ is the output of the GRU cell and will be passed forward.  The current hidden state $h_t$ is computed by:

\begin{linenomath}
\begin{equation}
h_t=u_t \odot h_{t-1}+(1-u_t) \odot \hat{h}_t
\end{equation}
\end{linenomath}

\subsubsection{Integrating GCN and GRU}
We integrate GCN and GRU to capture the spatial dependencies and temporal dependencies simultaneously. Let $A$ denote the adjacency matrix of spatial correlation graph $G$; let $X_t$ denote the historical travel demand. We first use GCN to process $A$ and $X_t$ using Equation~\ref{eq:z}:

\begin{equation} 
Z_t=f(X_t,A)=softmax(\widehat{A}~ReLU(\widehat{A}X_tW^0)~W^1)
\end{equation}

Then we concatenate the output matrix $Z_t$ with the temporal feature matrix $C_t$ to generate the input $Z_t' = [Z_t',C_t]$ of GRU. The calculation process in the GRU cell at time $t$ can be expressed as:

\begin{linenomath}
\begin{equation}
R_t=\sigma(Z_t'W_{ZR}+H_{t-1}W_{HR}+b_R)
\end{equation}
\begin{equation}
U_t=\sigma(Z_t'W_{ZU}+H_{t-1}W_{HU}+b_U)
\end{equation}
\begin{equation}
\widehat{H}_t=tanh(Z_t'W_{ZH}+(R_t \odot H_{t-1})W_{HH}+b_H)
\end{equation}
\begin{equation}
H_t=U_t \odot H_{t-1}+(1-U_t) \odot \widehat{H}_t
\end{equation}
\end{linenomath}\\
\noindent where $R_t$ is the reset gate, $U_t$ is the update gate, $H_{t-1}$ is the hidden state of the previous time step, $\widehat{H}_t$ is the candidate hidden state, $H_t$ is the current hidden state, $W_{ZR}$, $W_{HR}$, $W_{ZU}$, $W_{HU}$, $W_{ZH}$, $W_{HH}$ are weight parameters, and $b_R$, $b_U$, $b_H$ are biases.

\subsubsection{Constructing spatial correlation graph}
We use two graphs to represent the spatial correlations between areas.


\indent The first one is the environmental similarity graph $G_{F}=(V,E_{F})$ which is constructed by linking two environmentally similar areas $i$ and $j$ by an edge $e^{i,j}_F \in E_{F}$. The environmental similarity is evaluated by the same metrics (i.e., \textit{Pearson coefficient}) in \citet{tang2021multi}. In this study, we use the number of residential parcels, median parcel size (per race), high fire risk zone indicator and local responsibility area (LRA) indicator to estimate the environmental similarity. Let $A_{F}$ denote the adjacency matrix of graph $G_{F}$, the element $a_{F}^{i,j} \in A_{F}$ is given by:

\begin{linenomath}
\begin{equation}
a_{F}^{i,j}=
    \begin{cases}
    1, ~\text{if }sim(p_i,p_j)>d_F\\
    0, ~\text{otherwise.}
    \end{cases}
\end{equation}
\end{linenomath}\\
\noindent where $p_i \in \mathbb{R}^{1 \times n}$ and $p_j\in \mathbb{R}^{1 \times n}$ are the environmental feature vectors of nodes $i$ and $j$ respectively, $n$ is the number of features, $sim(\cdot)$ is the calculation function of the Pearson coefficient, and $d_F$ is the threshold parameter, which is set to 0.9 to control the sparsity of the graph. 

Similarly, the demographic similarity graph $G_{D}=(V,E_{D})$ is constructed by connecting two nodes $i$ and $j$ with similar demographic characteristics by an edge $e^{i,j}_D \in E_{D}$. Let $A_{D}$ denote the adjacency matrix of graph $G_{D}$, the element $a_{D}^{i,j} \in A_{D}$ is given by:

\begin{linenomath}
\begin{equation}
a_{D}^{i,j}=
    \begin{cases}
    1, ~\text{if }sim(q_i,q_j)>d_D\\
    0, ~\text{otherwise.}
    \end{cases}
\end{equation}
\end{linenomath}\\
where $q_i \in \mathbb{R}^{1 \times m}$ and $q_j\in \mathbb{R}^{1 \times m}$ are the demographic feature vectors of nodes $i$ and $j$ respectively, $m$ is the number of demographic features, $sim(\cdot)$ is the calculation function of the Pearson coefficient, and $d_D$ is the threshold parameter, which is also set to 0.9.

The environmental similarity and the demographic similarity are then fused to jointly represent the spatial correlations between areas. Specifically, the fused graph $G = (V, E)$ is established by connecting two areas $i$ and $j$ by an edge $e_{i,j} \in E$. Let $A$ denote the adjacency matrix of graph $G$, the element $a^{i,j} \in A$ is calculated by:

\begin{linenomath}
\begin{equation}
a^{i,j}=
    \begin{cases}
    1, ~\text{if }sim(p_i,p_j) + sim(q_i,q_j)>d_F + d_D\\
    0, ~\text{otherwise.}
    \end{cases}
\end{equation}
\end{linenomath}\\

In this way, the fused graph $G$ can largely maintain the sparcity and identify the areas with similar environmental and demographic characteristics.

\subsection{Model updating scheme}

Ideally, the GPS data providers may deliver near real-time GPS data with little delay. Hence, we may update our model, at least daily, to ensure that it can quickly adapt to the fast-changing situations during wildfires and thus better facilitate real-time decision-making. However, in real-world applications, data delivery is often delayed for at least 24 hours, due to the latency in data transfer, the time for data processing, and/or legal constraints. Therefore, this study considers both scenarios, with and without data delivery delay issues. 

\begin{figure}[H]
  \centering
  \includegraphics[width=0.98\textwidth]{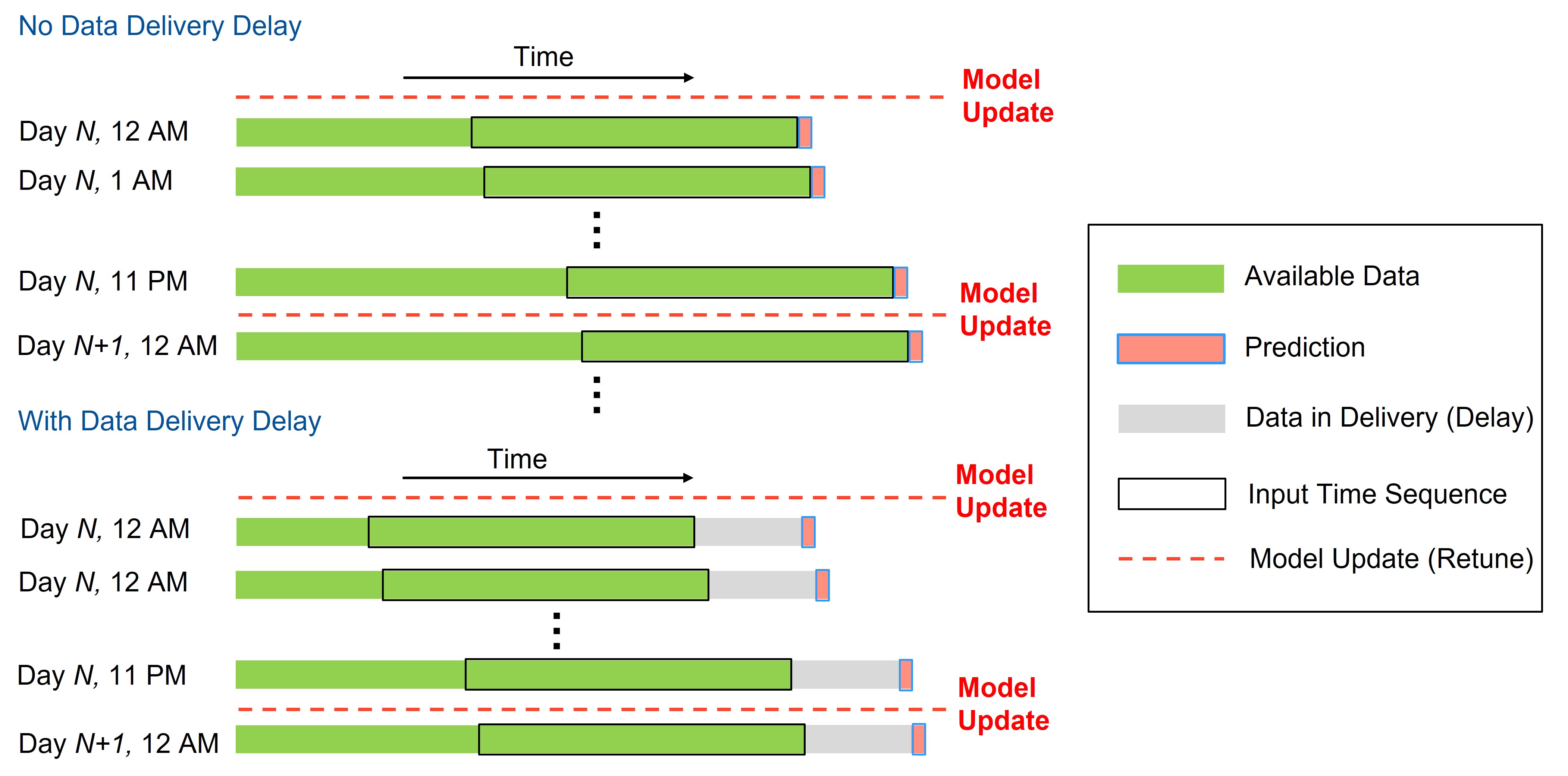}
  \caption{Model updating scheme considering data delivery delay}
  \label{fig:model_updating}
\end{figure}

Figure~\ref{fig:model_updating} illustrates the model updating scheme for both with and without delay. Suppose the GPS data is delivered on an hourly basis. The model takes a sequence of data as input (shown as the box) and forecasts the travel demand for the following hour (shown as the red bar). By the end of each day, we retrain/update the forecasting model with the available data. Then, we use the updated model to predict future travel demand on the following day. On the other hand, if the data delivery is delayed by multiple hours (as shown by the gray bar), we have to use less available data to train the model. 

For the case study (see Section~\ref{data}), since the wildfire started on October 23 and was fully contained on November 6, we stop updating the model after generating predictions for November 6 and update the model 14 times (i.e., 14-day duration of wildfire) in total. We consider three scenarios: 1) no delay, 2) 24-hour delay, and 3) 48-hour delay, and then compare the model's performance (see Section \ref{model comparison}). 




The model updating scheme enables the model to take in timely information as new data become available. During emergencies like wildfires, the situation is ever-changing, e.g., fire spread, weather, evacuation order/warning, among others, and the changing situation will greatly impact a householder's evacuation decision-making and travel behavior \citep{lindell2012protective}. Therefore, it is highly beneficial to recalibrate the forecasting model frequently in order to adapt to the rapidly changing situation and thus produce more accurate predictions to facilitate better-informed real-time decision-making.

\section{Data}
\label{data}

We selected the 2019 Kincade Fire, Sonoma County, CA, as the case study. This section introduces the study site and the wildfire in Subsection~\ref{ss:site} and describes the dataset in Subsection~\ref{ss:data}.

\subsection{Study site}
\label{ss:site}
\begin{figure}[H]
    \centering
    \includegraphics[width=\textwidth]{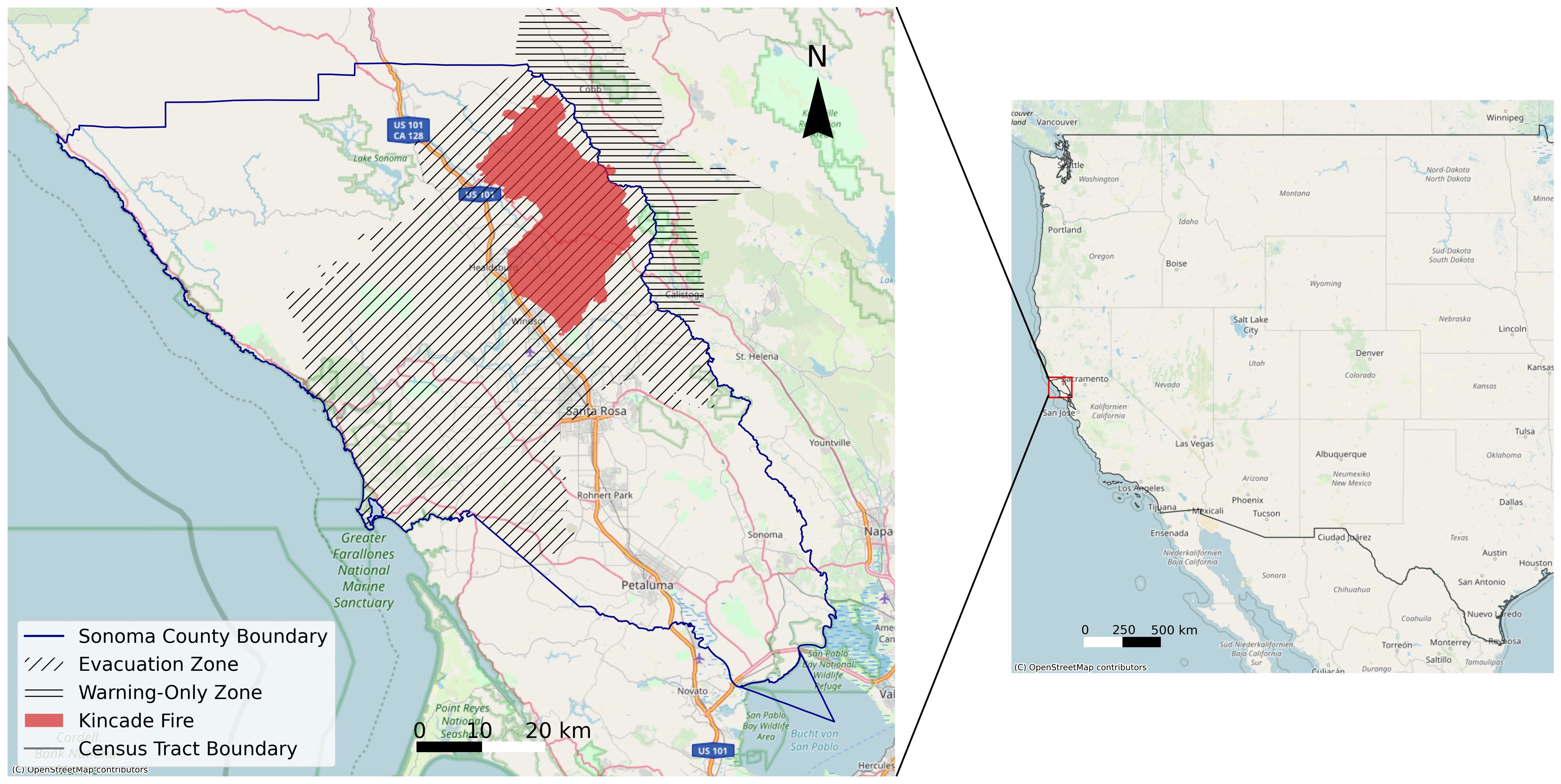}
    \caption{Sonoma County and the Kincade Fire perimeter}
    \label{fig:sonoma}
\end{figure}

Sonoma County is located in Northern California, U.S. The population estimate of Sonoma County was 494,336 in 2019. Its county seat and largest city is Santa Rosa. The highway system of Sonoma County consists of U.S. Highway 101, and State Highways 1, 12, 37, 116, 121, and 128. The Kincade Fire started northeast of Geyserville at 9:27 p.m. on October 23, 2019 and was fully contained at 7:00 p.m. on November 6, 2019. The fire burned 77,758 acres, destroyed 374 structures, damaged 60 structures, and caused 4 injuries \citep{sonoma2020report}. As the fire spread, the mandatory evacuation order was first issued in Geyserville at 10:00 a.m. on October 26, and then the evacuation warnings and orders grew to encompass nearly all of Sonoma County in the following days, making it the largest evacuation in Sonoma County's history. The study site and the fire perimeter are shown in Figure \ref{fig:sonoma}.

\subsection{Data description}
\label{ss:data}

The GPS data\footnote{The GPS data underwent Gravy's cleansing processes and was optimized with Gravy Location Data Forensics--filtering and categorizing inaccurate and even fraudulent location signals. This enabled us to identify and use only the cleansed location signals relevant to this project.} was provided by Gravy Analytics and built on privacy-friendly mobile location data. Gravy’s location data platform processes raw location signals from multiple data providers representing over 150 million U.S. mobile devices. The location information was aggregated into cells (the size is 4.77m $\times$ 4.77m) and represented by a 9-character geohash. The timestamp of the signal was encoded as epoch time in milliseconds rounded down to the floor minute. The accuracy of the signal location was indicated by the Forensic Flag (to measure the GPS error). The GPS error was usually caused by spoofed locations (e.g., some mobile applications may generate fake location information), IP address derived signals, Cell Tower Derived (CTD) signals, abnormal signals density, and GPS signal floating. In this study, we adopt the same data cleaning process as in \citet{zhao2022estimating}. Records with signal location error no more than 250 meters were retained. After the data cleaning process, we included 100,913,550 GPS signal records in Sonoma County, CA from October 16, 2019 to November 13, 2019 for analysis.

The hourly weather condition data were collected using the \textit{OpenWeather}\footnote{https://openweathermap.org/api} API. The parcel data (for functional similarity evaluation), the evacuation order and warning information, and the fire progression information were provided by the county authorities. The demographic data were collected from the \textit{American Community Survey 2015-2019 5-year} estimates data. Table~\ref{tab:variable} presents the descriptive statistics of the input variables used in the case study. The functional variables, the demographic variables, the fire distance, and the evacuation order/warning are processed at the census-tract level, and the weather condition variables are at the county level. In this study, the fire distance is defined as the shortest Euclidean distance between the fire perimeter and the boundary of each census tract. The evacuation order/warning is a dummy variable that indicates if a census tract is under mandatory evacuation order or warning at a given timestamp. The weekend indicator is also a dummy variable indicating weekday or weekend. We also included historical travel demand information, i.e., historical embeddings, as two additional inputs. Specifically, historical embeddings 1 refers to the overall historical average travel demand for each census tract before given timestamp while historical embeddings 2 refers to the only last-four-hour historical average travel demand for each census tract. Finally, we calculate the population change (i.e., the number of active users on each day after wildfire divided by the average number of active users per weekday before wildfire) to capture the population movement during wildfire evacuation.


\begin{table}[]
\centering
\caption{Descriptive statistics of input variables}
\label{tab:variable}
\resizebox{\textwidth}{!}{%
\begin{tabular}{@{}llllll@{}}
\toprule
Variables                                           & Mean      & Std.      & Min       & Max        & Category                                                                                      \\ \midrule
Residential area count                              & 1653.091  & 713.102   & 227.000   & 4333.000   & \multirow{4}{*}{\begin{tabular}[c]{@{}l@{}}Environmental \\ similarity \\ graph\end{tabular}} \\
Median parcel size (per arce)                       & 0.518     & 1.308     & 0.033     & 9.355      &                                                                                               \\
High fire risk zone                                 & 0.152     & 0.360     & 0.000     & 1.000      &                                                                                               \\
If the zone is inside LRA                           & 0.697     & 0.462     & 0.000     & 1.000      &                                                                                               \\ \midrule
Population density (per sq. mile)                   & 3339.788  & 3192.117  & 7.155     & 12474.630  & \multirow{6}{*}{\begin{tabular}[c]{@{}l@{}}Demographic \\ similarity \\ graph\end{tabular}}   \\
Proportion of the white population                  & 0.766     & 0.129     & 0.382     & 0.955      &                                                                                               \\
Proportion of population with BA’s degree and above & 0.363     & 0.127     & 0.121     & 0.636      &                                                                                               \\
Median household income (US dollar)                 & 83823.667 & 20522.564 & 49856.000 & 145147.000 &                                                                                               \\
Proportion of households own 0 car                  & 0.075     & 0.068     & 0.000     & 0.320      &                                                                                               \\
Median age                                          & 43.801    & 8.960     & 23.000    & 71.600     &                                                                                               \\ \midrule
Fire distance                                       & 181.731   & 102.993   & 0.000     & 250.000    & \multirow{13}{*}{\begin{tabular}[c]{@{}l@{}}Temporal \\ variables\end{tabular}}               \\
Evacuation order/warning                            & 0.101     & 0.301     & 0.000     & 1.000      &                                                                                               \\
Weekend indicator                                   & 0.308     & 0.462     & 0.000     & 1.000      &                                                                                               \\
Temperature                                         & 56.762    & 12.803    & 33.400    & 90.600     &                                                                                               \\
Feels like temperature                              & 56.521    & 12.574    & 33.400    & 86.400     &                                                                                               \\
Wind speed                                          & 4.017     & 4.314     & 0.000     & 29.600     &                                                                                               \\
Sea level pressure                                  & 1016.715  & 3.230     & 1005.900  & 1023.900   &                                                                                               \\
Humidity                                            & 58.437    & 29.248    & 8.920     & 100.000    &                                                                                               \\
Visibility                                          & 8.752     & 2.416     & 0.000     & 9.900      &                                                                                               \\
Cloud cover                                         & 17.321    & 26.958    & 0.000     & 100.000    &                                                                                               \\
UV index                                            & 1.535     & 2.201     & 0.000     & 8.000      &                                                                                               \\
Historical embeddings 1                             & 204.004   & 124.704   & 0.000     & 1210.731   &                                                                                               \\
Historical embeddings 2                             & 184.727   & 127.465   & 0.000     & 1284.838   &                                                                                               \\
Population Change                                   & 0.819     & 0.225     & 0.000     & 1.706      &                                                                                               \\ \bottomrule
\end{tabular}%
}
\end{table}

\section{Results}
\label{results}

In the following section, we will sequentially introduce the results of the inferred trip generations, the implementation settings of SA-MGCRN, the brief profile of all benchmark models, the prediction performance of SA-MGCRN and benchmark models across three delay scenarios and the predictive contributions of the major components embedded in SA-MGCRN via an ablation study.

\subsection{Trip generation inference}
\begin{figure}[H]
    \centering
    \includegraphics[width=0.75\textwidth]{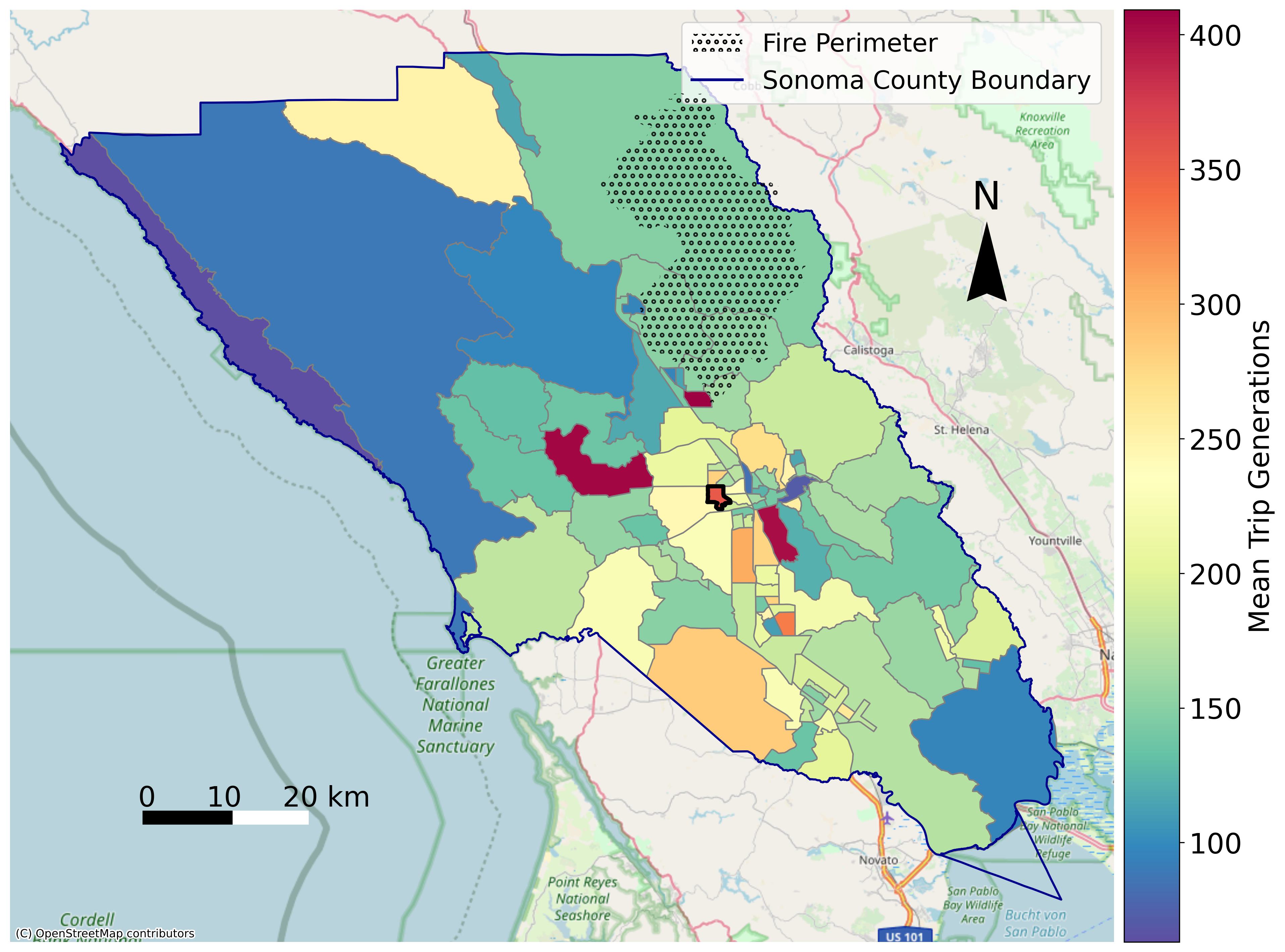}
    \caption{Spatial distribution of mean trip generation in Sonoma County (census-tract level).}
    \label{fig:dist}
\end{figure}

We inferred 422,466 trips generated in Sonoma County from October 18, 2019 to November 13, 2019 using the GPS data. Then, we aggregated these trip generations into the census-tract level and count the hourly travel demand for each census tract. We further infer the trip generations (i.e., travel demand) for the whole population following the procedure described in Section \ref{trip inference}. The spatial distribution of trip generations is presented in Figure~\ref{fig:dist}. The mean of hourly inferred travel demand for each census tract is 184.647, the standard deviation is 152.719, the maximum value is 1769.286, and the minimum value is 0. The mean of the number of active users per census tract is 168.402, the standard deviation is 106.993, the maximum value is 564.25, and the minimum value is 29.5.

\subsection{Model setting}
The case studies were conducted with Adam optimizer in a Pytorch environment \citep{paszke2019pytorch} using an Ampere A-100 GPU. We tuned the hyperparameters, including batch size, sequence length and learning rate, of SA-MGCRN model for each instance (i.e., each date and each data delivery data scenario). We adopted an early-stopping technique to prevent the model from overfitting and improve the training efficiency. If the trained model does not have an accuracy improvement during the last 200 training epochs, we consider the model has already converged. In our case study, we update (i.e., retune) SA-MGCRN 14 times (from October 23, 2019 to November 6, 2019) in total. The testing set is set as the travel demand on the following day after each model updates. The validation set is set as the travel demand of the day before the testing set day and the training set is set as the travel demand data before validating set day.

\subsection{Model comparison}
\label{model comparison}
In this section, we compared the proposed SA-MGCRN model with several benchmark models. The details of these models are described as follows. Note that all the models were fine-tuned.
\begin{itemize}
    \item \textbf{HA}: Historical Average is one of the most fundamental statistical models for time series prediction. HA predicts the demand in a specific timestamp period by averaging historical observations \citep{hyndman2018forecasting}. 
    \item \textbf{ARIMA}: Auto-Regressive Integrated Moving Average is a statistical time series prediction model. ARIMA fits a parametric model based on historical observations to predict future demand. The order of ARIMA was set to (3,1,1) in the case study.
    \item \textbf{SVR}: Support Vector Regression is a machine learning model that uses the same principle as Support Vector Regression (SVM) but for regression problems. We used the Radial Basis Function (RBF) kernel here. The cost was set to 10.
    \item \textbf{GBDT}: Gradient Boosting Decision Tree is a tree-based ensemble machine learning model. In this case study, the number of trees was set to 300, the maximum depth was set to 5, and the learning rate was set to 0.05. 
    \item \textbf{RF}: Random Forest is another tree-based ensemble machine learning method. In this model, the number of trees was set to 100, and we consider all features when looking for the best split.
    \item \textbf{MLP}: Multiple Layer Perceptron is a classical feedforward artificial neural network. In the case study, we used an MLP model with 64 neurons in one hidden layer, followed by a dropout layer to prevent the model from overfitting. Both the learning rate, batch size, the sequence length are fine-tuned at each trial.
    \item \textbf{LSTM}: Long Short-Term Memory is another widely used neural network based on the gating mechanism. Both the learning rate, batch size, the sequence length are fine-tuned for each trail.
\end{itemize}

We compared and evaluated the prediction performance of all the benchmarks and SA-MGCRN using Mean Absolute Error (MAE) and Root Mean Square Error (RMSE). These metrics can be defined as:

\begin{linenomath}
\begin{equation}
MAE = \frac{1}{n}\sum_{i=1}^{n}\left | y_{i} - \hat{y_{i}} \right |,
\end{equation}
\begin{equation}
RMSE = \sqrt{\frac{1}{n}\sum_{i=1}^{n}(y_{i} - \hat{y_{i}})^{2}},
\end{equation}
\end{linenomath}\\
where $y$ are the observed trip generations and $\hat{y}$ are the predicted trip generations.
\begin{table}[]
\centering
\caption{Prediction performance of SA-MGCRN and benchmark models across all census tracts in Sonoma County}
\label{tab:performance}
\resizebox{0.9\textwidth}{!}{%
\begin{tabular}{@{}lllllll@{}}
\toprule
\multirow{2}{*}{Models} & Delay = 0       & \multicolumn{1}{l|}{}     & Delay = 24h     & \multicolumn{1}{l|}{}     & Delay = 48h     &                  \\ \cmidrule(l){2-7} 
                        & MAE             & \multicolumn{1}{l|}{RMSE} & MAE             & \multicolumn{1}{l|}{RMSE} & MAE             & RMSE             \\ \midrule
HA                      & 97.439          & 128.509                   & 99.933          & 131.622                   & 101.984         & 133.960          \\
ARIMA                   & 109.7043        & 150.82                    & 111.105        & 151.381                  & 113.250        & 154.481         \\
SVR                     & 87.197          & 117.177                   & 91.196          & 121.975                   & 93.248          & 125.185          \\
RF                      & 86.033          & 116.703                   & 90.593          & 122.671                   & 93.903          & 126.932          \\
GBDT                    & 90.770          & 124.246                   & 96.278          & 132.083                   & 100.234         & 137.439          \\
MLP                     & 82.287          & 112.629                   & 87.339          & 118.814                   & 90.446          & 123.195          \\
LSTM                    & 89.637          & 119.010                   & 92.915          & 122.805                   & 95.686          & 126.075          \\
SA-MGCRN                & \textbf{78.616} & \textbf{108.121}          & \textbf{83.378} & \textbf{114.049}          & \textbf{89.068} & \textbf{121.015} \\ \bottomrule
\end{tabular}%
}
\end{table}

\begin{figure}[H]
    \centering
    \includegraphics[width=0.95\textwidth]{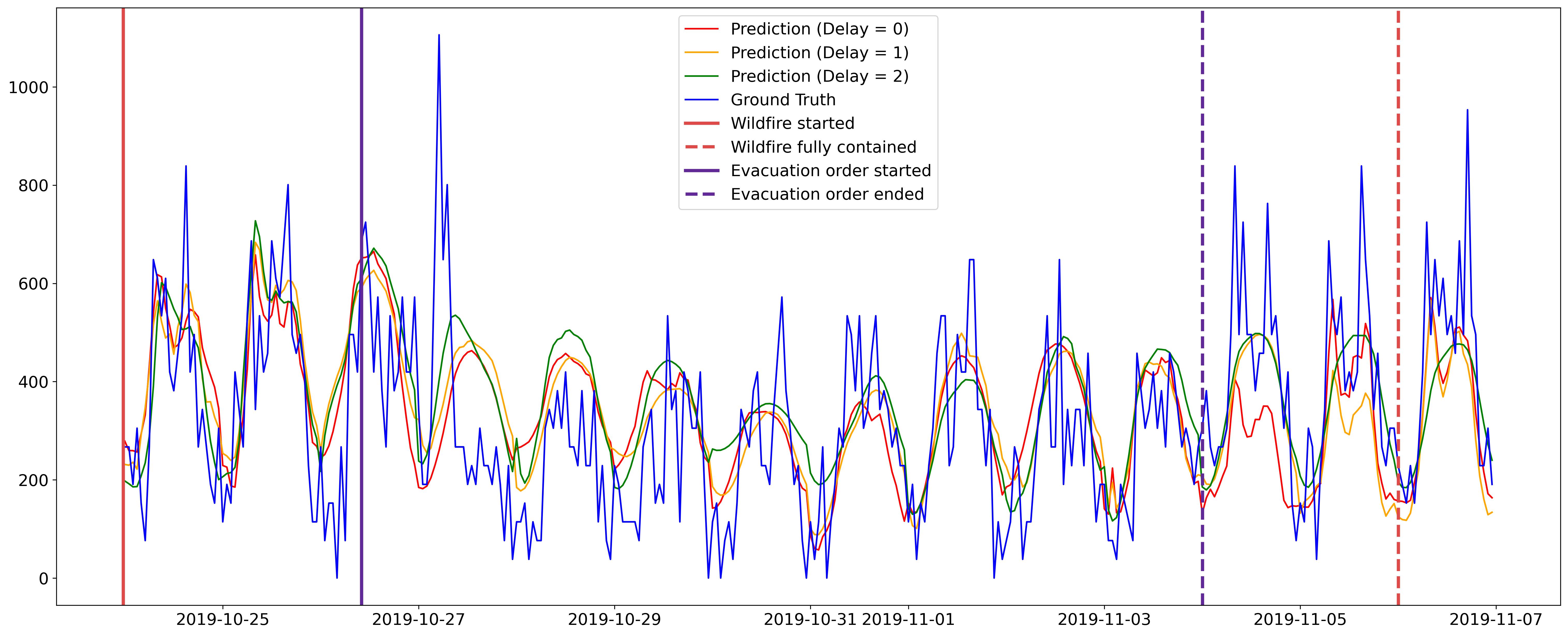}
    \caption{Hourly prediction performance of SA-MGCRN considering data delivery delay for a selected census tract. Please refer to the census tract with highlighted boundary in Figure \ref{fig:dist}.}
    \label{fig:comparison}
\end{figure}

The model performance results across all areas are shown in Table~\ref{tab:performance}. First, when there are no data delivery delay issues, the SA-MGCRN model significantly outperforms all the benchmark models. MLP shows better prediction performance than the statistical models (i.e., HA and ARIMA) and the classical machine learning models (i.e., SVR, GBDT, and RF). Among the classical machine learning models, the RF model had the best performance. LSTM delivers a relatively lower predictive performance, probably because its complex model structure is not appropriately applicable to the small-size training set ($n$ = 624). The prediction accuracy across all models showcases a decreasing trend when the days of delay gradually increase. For example, MAE of SA-MGCRN increases from 78.616 to 83.378 and 89.068 when there are 24 and 48 hours delay. This is intuitive since more delay usually leads to the absent awareness of the ongoing (wildfire) situation, which inevitably restricts the model's prediction performance. Notably, our proposed SA-MGCRN model is still the best-performing one among all models even if there exist data delivery delay issues regarding MAE and RMSE. We also illustrated the hourly prediction performance of SA-MGCRN during the wildfire (i.e., October 24 to November 6, 2019) of a specific census tract in Figure~\ref{fig:comparison}. The results show that the prediction accuracy of the SA-MGCRN model can well pick up the temporal trend of the trip generation rate under different levels of delay. However, SA-MGCRN can hardly address the extreme values.  Figure~\ref{fig:comparison} also shows that the highest predictive error occurs on October 27. This is probably because wildfire-related information (e.g., evacuation order and fire distance) is lacking until the wildfire started (after October 26). The predictive error significantly drops as the model keeps updating itself by absorbing wildfire-related information.

\begin{table}[]
\centering
\caption{Prediction performance of SA-MGCRN and benchmark models across areas with evacuation orders or warnings in Sonoma County}
\label{tab:performance_ordered}
\resizebox{0.9\textwidth}{!}{%
\begin{tabular}{@{}lllllll@{}}
\toprule
              & Delay = 0       & \multicolumn{1}{l|}{}     & Delay = 24h     & \multicolumn{1}{l|}{}     & Delay = 48h     &                  \\ \cmidrule(l){2-7} 
              & MAE             & \multicolumn{1}{l|}{RMSE} & MAE             & \multicolumn{1}{l|}{RMSE} & MAE             & RMSE             \\ \midrule
HA            & 94.756          & 126.645                   & 98.569          & 131.956                   & 101.753         & 135.784          \\
ARIMA         & 108.833         & 152.599                   & 111.370         & 153.892                   & 113.950         & 157.750          \\
SVM           & 90.895          & 122.350                   & 96.717          & 129.141                   & 99.577          & 133.510          \\
RF            & 90.797          & 123.338                   & 97.239          & 131.396                   & 102.290         & 137.585          \\
GBDT          & 95.564          & 131.815                   & 102.848         & 141.525                   & 108.704         & 149.429          \\
MLP           & 85.642          & 117.864                   & 93.253          & 127.028                   & 96.993          & 132.511          \\
LSTM          & 90.575          & 121.115                   & 94.563          & 125.517                   & 97.599          & \textbf{128.356} \\
\textbf{SA-MGCRN} & \textbf{80.555} & \textbf{111.731}          & \textbf{86.598} & \textbf{119.151}          & \textbf{94.766} & 128.791          \\ \bottomrule
\end{tabular}%
}
\end{table}

We also compare the prediction performance of SA-MGCRN and benchmark models across areas with evacuation orders or warnings in Sonoma County (please refer to Figure \ref{fig:sonoma} for the spatial distribution of these areas), as shown in Table~\ref{tab:performance_ordered}. The results show that all models have relatively lower prediction performance than the performance across all areas shown in Table \ref{tab:performance}. However, our proposed model, i.e., SA-MGCRN, is still the best-performing model in no delay, 24-hour delay, and 48-hour delay scenarios\footnote{LSTM has slightly better RMSE than SA-MGCRN in the 48-hour delay scenario, but the difference in RMSE between the two models is negligible ($\sim$0.3\%).}.

\subsection{Ablation study}
We conducted an ablation study for SA-MGCRN with no data delivery delay. The ablation study examines the performance of the model by removing certain components to see the contribution of the removed components \citep{meyes2019ablation}. The component is considered to be more predictively important if removing itself increases the model error since in this case the model relatively depends more on this component for the predictions. In this ablation study, we generate nine models by sequentially removing environmental similarity graph, demographic similarity graph, weekend indicator, weather information, evacuation order/warning information, fire distance, two historical embeddings and population change.

\begin{table}[H]
\centering
\caption{Results of ablation study}
\label{tab:ablation}
\resizebox{\textwidth}{!}{%
\begin{tabular}{@{}lllll@{}}
\toprule
Components                            & MAE             & Importance \% (by MAE) & RMSE             & Importance \% (by RMSE) \\ \midrule
\textbf{SA-MGCRN}                   & \textbf{78.616} & -                      & \textbf{108.121} & -                       \\
W/O environment similarity            & 78.675          & 0.397                  & 108.303          & 1.296                   \\
W/O demographic similarity            & 80.157          & 10.327                 & 109.726          & 11.447                  \\
W/O weekend indicator                 & 79.612          & 6.678                  & 108.830          & 5.060                   \\
W/O weather information               & 79.673          & 7.082                  & 109.035          & 6.517                   \\
\textbf{W/O evacuation order/warning} & \textbf{81.592} & \textbf{19.954}        & \textbf{111.773} & \textbf{26.052}         \\
\textbf{W/O fire distance}            & \textbf{80.969} & \textbf{15.772}        & \textbf{110.094} & \textbf{14.071}         \\
W/O historical embeddings 1           & 80.408          & 12.011                 & 109.787          & 11.880                  \\
W/O historical embeddings 2           & 80.410          & 12.030                 & 109.385          & 9.014                   \\
\textbf{W/O population change}        & \textbf{80.965} & \textbf{15.749}        & \textbf{110.177} & \textbf{14.662}         \\ \bottomrule
\end{tabular}%
}

\parbox[t]{0.98\textwidth}{\vskip3pt{\footnotesize Notes: Historical Embeddings 1 is calculated as the entire historical average at the given timestamp; Historical Embedding 2 is calculated as the last four hour average. The importance of the $k$-th component is mathematically calculated as $ e_k \times 100\% / \sum_{i=1}^{n} e_i $, where $e_k$ refers to the increased predictive error after removing the $k$-th component. }}

\end{table}

The performance of the nine ablated models is presented in Table~\ref{tab:ablation}. We calculate the importance of each component by two performance metrics, i.e., MAE and RMSE, to ensure the robustness of the results. The results suggest that all components contribute to the prediction accuracy, with the evacuation order/warning information, fire distance and popotation change being the top 3. Removing wildfire-related information, i.e., evacuation order/warning, showcases considerable predictive power. Results demonstrate that it accounts for 19.954\% and 26.052\% of importance calculated by MAE and RMSE. Removing fire distance also notably increases the prediction error. Specifically, the predictive importance of fire distance calculated by MAE and RMSE is 15.772\% and 14.071\%, respectively. According to the results, the prediction error considerably increases compared with the original model when removing the population change (MAE increases 15.749\% and RMSE increases 14.662\%). A relatively small increase in prediction error (less than 1\% for MAE and 1.296\% for RMSE) occurs when we remove the environmental similarity graph. This finding indicates that although environmental similarity graph can improve the model performance, the contribution is relatively limited. Instead, demographic simiarity graph shows greater contribution to the predictive performance. Removing this graph yields a 10.327\% and 11.447\% increase of MAE and RMSE. Weather information accounts for a relatively small strength in terms of forecasting travel demand (7.082\% for MAE and 6.517\% for RMSE). Two historical embeddings (i.e., historical travel demand information) display similar contributions in predicting travel demand under wildfire evacuations. Specifically, historical embeddings 1, i.e., the entire historical average of travel demand only accounts for around 12\% of predictive importance. Historical embeddings 2, i.e., the last-four-hour average of travel demand, contributes 12.030\% and 9.014\% of predictive importance calculated by MAE and RMSE.

\section{Discussion and conclusion}
\label{discussion and conclusion}

In this study, we develop a Situational-Aware Multiple Graph Convolutional Recurrent Network (SA-MGCRN) model to forecast the real-time travel demand in wildfire evacuations while accounting for data delivery delay issues. The proposed model incorporates multiple dimensions of information including historical census-tract-level travel demand, temporal features (i.e., weather, evacuation order/warning, fire progression, weekend indicator and population change), and spatial features (i.e., socio-demographics similarity graph and built environment similarity graph). The proposed model, which is based on the framework in Figure~\ref{fig:pred_frmk}, uses these inputs to make accurate predictions on an hourly basis for the following day. Given the difficulties of having reliable traffic data during wildfire emergencies \citep{melendez2021modelling}, this study utilizes GPS data collected from mobile devices to derive trips (for daily travel or evacuation purposes) for each census tract using the clustering approach illustrated in Figure~\ref{fig:clustering}.

Using GPS data for real-time trip generation forecasting in wildfire scenarios is a novel and promising approach. This type of information provides emergency managers and planners with critical and reliable information on what is happening in the areas of interest and facilitates them to decide where and which type of emergency management strategies (e.g., traffic signal retiming, extending the area under evacuation order or activating contraflow) should be prioritized to keep mobility under emergent scenarios. Furthermore, the proposed SA-MGCRN model can provide them with accurate simulations (i.e., predictions) regarding the pattern of travel demand in the following day and therefore help pre-evaluate the effectiveness of the possible decisions. As such, this work takes a substantial step forward in the literature on the use of GPS data for wildfire evacuation management. To date, GPS historical data have been mostly used only to investigate wildfire evacuation after several months from the events \citep{zhao2022estimating}. Therefore, this work represents the first attempt to use GPS big data and AI to provide near real-time forecasting of human behavior to emergency managers and evacuation modelers during a wildfire disaster. In fact, the output of the SA-MGCRN can be used as an input for existing wildfire evacuation models for near real-time evacuation simulations of different if-then scenarios.

SA-MGCRN shows strong capabilities in coping with spatial and temporal dependencies. Specifically, SA-MGCRN first employs GCN model to account for spatial correlations between census tracts using several graphs (i.e. environmental similarity graph and demographic similarity graph). Then, a GRU model is integrated to extract temporal dependencies behind multiple temporal variables including historical demand, weather, evacuation order/warning information, fire progression, weekend indicator and population change. The case study of 2019 Kincade Fire shows that SA-MGCRN outperforms all state-of-the-art benchmark models (see Table~\ref{tab:performance}, Table~\ref{tab:performance_ordered} and Figure~\ref{fig:comparison}). In real-world applications, the GPS data usually suffer delivery delay issues, which may impede the real-time decision-making process. This study accounts for this issue by considering both delay and non-delay scenarios. Results showed that more data delivery delays will trigger an accuracy drop and our proposed model, i.e., SA-MGCRN, is still the best-performing model among both statistical and machine learning models. We conducted an ablation study to quantitatively investigate the contribution of each component on prediction accuracy. Results (in Table~\ref{tab:ablation}) highlight that the most important component is the evacuation order/warning information, followed by fire distance and population change. 

Previous studies have highlighted the critical role of social cues and crisis communication in determining evacuation behavior \citep{kuligowski2022modeling, lovreglio2019modelling}. Our study further reinforces this finding by demonstrating that the evacuation order and warning information is the most significant predictor for accuracy in forecasting travel demand under wildfire evacuation. In addition, this work showcases the strong relevance of the fire distance when forecasting traffic demand during wildfire evacuations, which aligns with a recent study \citep{wu2022wildfire}. While another previous work by \citet{lovreglio2020calibrating} did not show evidence that this variable was significant in the decision to evacuate. This can be possibly attributed to the difference in the adopted approaches. \citet{lovreglio2020calibrating} used a statistical model to assess the impact of the distance between the households and the \textit{final} fire front. However, this study uses a deep learning approach and considers that the fire distance changes dynamically on a \textit{daily} basis depending on the new fire front location. We believe that fire distance changes quickly throughout the fire progression and naturally influences the household's decision to evacuate or not. Our proposed model can well capture these spatiotemporal dynamics and therefore highly relies on this component. 

Overall, our findings show that the proposed methodological framework can generate highly accurate predictions to facilitate real-time emergency management. However, there are several limitations that require follow-up work. In this study, fire cues were accounted for just by using the distance from the fire perimeter. However, wildfire generates smoke and embers, which can reach areas before the fire front reaches them, depending on the weather conditions \citep{kochanski2019modeling, ronchi2017sanctuary}. Also, this study did not include social influence as a component of SA-MGCRN, as capturing social interactions by solely using large-scale GPS data can be challenging. In fact, social influence can play a crucial role in determining evacuation behavior as people may respond when others are responding \citep{lovreglio2016evacuation, reneke2013evacuation}. In future work, innovative solutions need to be developed to include social influence characteristics into the current modeling framework to further improve prediction accuracy. Additionally, the fire distance is measured by the distance from the centroid of each census tract to the fire front instead of the actual distance from a householder to the fire front. This may create potential uncertainties. Figure \ref{fig:comparison} shows that SA-MGCRN has relatively limited capabilities to deal with extreme values. One possible explanation is that the size of the training sample is relatively small ($n$ = 624). We believe that with more observations, SA-MGNRN can further address the extreme values and produces even higher prediction accuracy. Notably, SA-MGCRN has the worst predictive performance at the start of the wildfire, due to limited previous wildfire-related information for training. To address this issue, future studies may consider integrating human intelligence into machine learning (i.e., human-in-the-loop AI \citep{monarch2021human}) to further reduce the prediction error \citep{zhang2022machine}. Furthermore, more variables such as point of interest information can be included in the model as they are also relevant to travel demand prediction \citep{xu2023real}. Another limitation is that we only use one case study to evaluate our proposed model's performance and the results may not be directly transferable to other contexts \citep{pel2012review}. Future studies may consider assessing the transferability of our proposed model by including more wildfire case studies.

\section*{Authorship Contribution Statement}
The authors confirm contributions to the paper as follows: \textbf{Zhang}: Conceptualization, Data Curation, Methodology, Software, Formal Analysis, and Draft Preparation. \textbf{Zhao}: Conceptualization, Methodology, Formal Analysis, Draft Preparation, Supervision and Grant Acquisition. \textbf{Xu}: Methodology, Data Curation, Visualization, Draft Preparation. \textbf{Lovreglio}: Conceptualization, Formal Analysis, Draft Preparation. \textbf{Nilsson}: Conceptualization, Formal Analysis. 

\section*{Acknowledgment}
This work was performed under the following financial assistance award No. 60NANB20D182, 60NANB21D180, and 60NANB22D177 from U.S. Department of Commerce, National Institute of Standards and Technology (NIST). Any opinions, findings, conclusions, or recommendations expressed in this material are those of the authors and do not necessarily reflect the views of NIST. We would like to thank Ruoyang Xiong for his contributions to the previous draft of the paper.
\newpage

\FloatBarrier


\FloatBarrier
\newpage





\bibliographystyle{elsarticle-harv}
\biboptions{semicolon,round,sort,authoryear}
\bibliography{sample.bib}







\end{document}